\documentclass[conference]{IEEEtran}
\IEEEoverridecommandlockouts
\usepackage{cite}
\usepackage{amsmath,amssymb,amsfonts}
\usepackage{algorithmic}
\usepackage{graphicx}
\usepackage{textcomp}
\usepackage{xcolor}
\def\BibTeX{{\rm B\kern-.05em{\sc i\kern-.025em b}\kern-.08em
    T\kern-.1667em\lower.7ex\hbox{E}\kern-.125emX}}

\usepackage{booktabs} % Francisco    
\usepackage{subfig} % Francisco
\usepackage{multirow} % Francisco

\begin{document}
\title{Boosting Kidney Stone Identification in Endoscopic Images Using Two-Step Transfer Learning %*\\
}

%%%%%%%%%%%%%%%%%%%%%%%%%%%%%%%%%%%%%%%%%%%%%%%%%%%%%%%%%%%%%%%%%%%%%%%%%%%%%%%%%%%%%%%%%%%%%

\author{Francisco Lopez-Tiro$^{1,2,3}$, Juan Pablo Betancur-Rengifo$^{3}$, Arturo Ruiz-Sanchez$^{3}$, Ivan Reyes-Amezcua$^{3,4}$, \\ Jonathan El-Beze$^{5}$, Jacques Hubert$^{5}$, Michel Daudon$^{6}$, Gilberto Ochoa-Ruiz*$^{,1,3}$, Christian Daul*$^{,2}$ \\

%Francisco Lopez-Tiro, Juan Pablo Betancur-Rengifo, Arturo Ruiz-Sanchez, Ivan Reyes-Amezcua, Jonathan El-Beze, Jacques Hubert, Michel Daudon, Gilberto Ochoa-Ruiz, Christian Daul

\thanks{$^{1}$Tecnologico de Monterrey, School of Sciences and Engineering, Mexico}%
\thanks{$^{2}$CRAN (UMR 7039, Université de Lorraine and CNRS), Nancy, France}%
\thanks{$^{3}$CV-INSIDE Lab Member, Mexico}
\thanks{$^{4}$CINVESTAV, Computer Sciences Department, Mexico}%
\thanks{$^{5}$CHU Nancy, Service d’urologie de Brabois, Vand{\oe}uvre-l\`es-Nancy, France }%
\thanks{$^{6}$H\^opital Tenon, AP-HP, Paris, France}%

\thanks{*Corresponding authors:} %
\thanks{gilberto.ochoa@tec.mx, christian.daul@univ-lorraine.fr}
}

\maketitle

\begin{abstract}

Knowing the cause of kidney stone formation is crucial to establish treatments that prevent recurrence. There are currently different approaches for determining the kidney stone type. However, the reference ex-vivo identification procedure can take up to several weeks, while an in-vivo visual recognition requires highly trained specialists. Machine learning models have been developed to provide urologists with  an automated classification of kidney stones during an ureteroscopy; however, there is a general lack in terms of quality of the training data and methods. In this work, a two-step transfer learning approach is used to train the kidney stone classifier. The proposed approach transfers knowledge learned on a set of images of kidney stones acquired with a CCD camera (ex-vivo dataset) to a final model that classifies images from endoscopic images (ex-vivo dataset). The results show that learning features from different domains with similar information helps to improve the performance of a model that performs classification in real conditions (for instance, uncontrolled lighting conditions and blur). Finally, in comparison to models that are trained from scratch or by initializing ImageNet weights, the obtained results suggest that the two-step approach extracts features improving the identification of kidney stones in endoscopic images.
\end{abstract}

\begin{IEEEkeywords}
Transfer learning, kidney stones, deep learning
\end{IEEEkeywords}

\section{Introduction} % Francisco
\vspace{-0.1cm}

The formation of kidney stones in the urinary tract is a public health issue \cite{hall2009nephrolithiasis, kasidas2004renal}. In industrialized countries, 10\% of the population suffers from an episode of kidney stones during their lifetime. Recent studies have determined that the risk of recurrence increases up to $40\%$ in less than 5 years \cite{kartha2013impact, scales2012prevalence}.
Thus, determining the root cause of kidney stone formation is crucial to avoid relapses through personalized treatments \cite{friedlander2015diet, kartha2013impact, viljoen2019renal}. 
Therefore, different approaches for visually identifying some of the most common types (or classes) of kidney stones have been proposed in recent years \cite{daudon2004clinical, estrade2017should}. 

The Morpho-Constitutional Analysis (MCA) is currently  the reference method for the identification of the type of the extracted kidney stone fragments \cite{corrales2021classification}. This ex-vivo procedure consists of two complementary analyses on the extracted kidney stone parts, which were fragmented with a laser. The fragments are visually inspected under a microscope to observe the colors and textures  of their surface and section. 
Then, an infrared-spectrophotometry analysis enables to identify the molecular and crystalline composition of the different areas (layers) of the kidney stone \cite{daudon2016comprehensive}.
However, in numerous hospitals the MCA results are only available after some weeks. This delay makes it difficult to establish an immediate and appropriate treatment for the patient. On the other hand, removing large kidney stone fragments is often difficult in practice. 
%% Not true : ! To complicate the matters, in modern ureteroscopic procedures, kidney stones are pulverized with a laser during surgery (i.e., laser dusting). 
Moreover, the biochemical composition can be altered by the laser during the fragmentation \cite{keller2019fragments}, making the MCA procedure challenging in some cases.

Endoscopic Stone Recognition (ESR) is a promising technique to immediately determine the type of kidney stones during the ureteroscopy (i.e., in-vivo recognition). 
The advantage of ESR is twice: kidney stones can be pulverized (dusting procedure with a laser) instead fragmented, and an appropriate treatment can be immediately defined.    
ESR is only based on a visual inspection  of in-vivo endoscopic images observed on a screen. For trained urologists, ESR results are strongly correlated with those of MCA  \cite{estrade2022towards}. However, only a few highly trained experts are nowadays able to recognize the type of kidney stones using only endoscopic images. Moreover, the visual classification by urologists is operator dependent and subjective, and the required experience is long to acquire   \cite{de2019metabolic}.%, krambeck2010analysis}. 

Studies have been recently proposed to automate ESR \cite{black2020deep, estrade2022towards, lopez2021assessing}. These Deep Learning (DL) based methods led to promising results. However, one of the most common challenges in these DL-based methods for classifying kidney stones is the lack of a large image set for the model training. In addition, the similarity of the data distribution is another important factor to obtain an adequate model. Consequently, this suggests a trade-off between the amount of available data and the data distribution to fit the network weights adequately. The majority of the DL-based models report fine-tuning of weights learned from distributions other than those from kidney stone images (commonly from ImageNet \cite{deng2009imagenet}). 

\begin{figure*} [] 
    \centering
    \subfloat[Dataset A: CCD-camera images (ex-vivo) ]{
    \label{fig:dataseta}\includegraphics[width=0.33\linewidth]{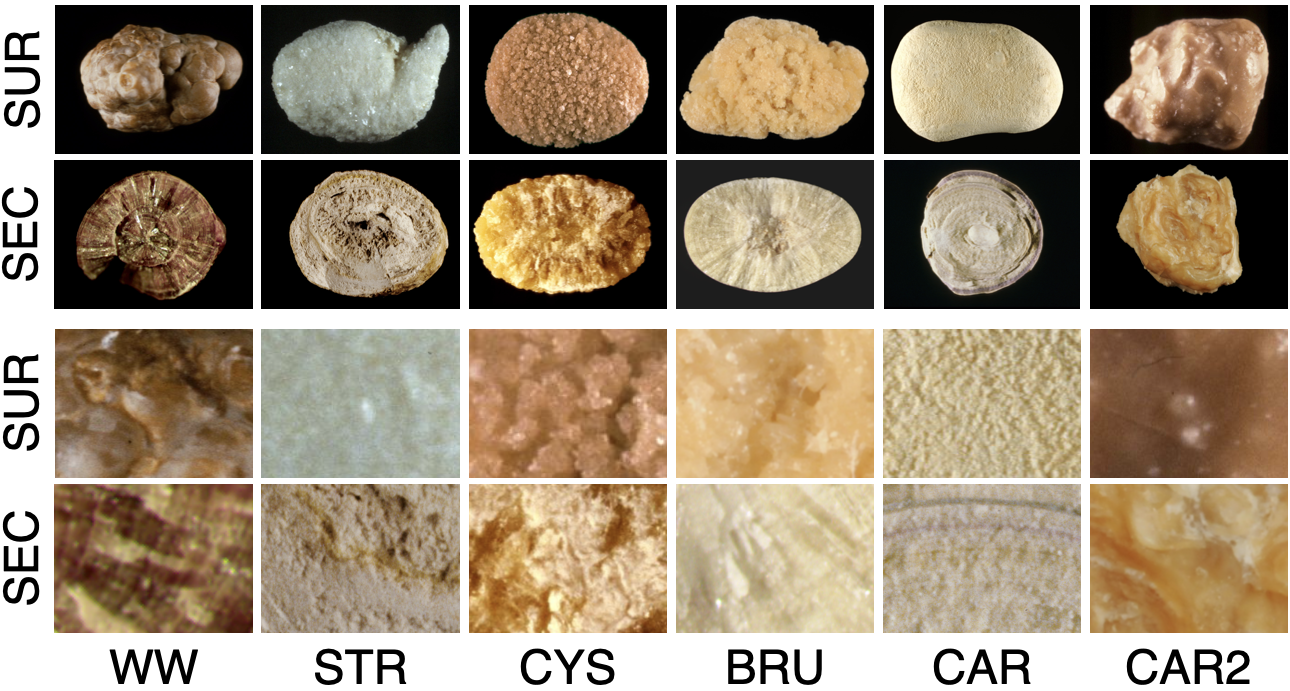}}
    \hspace{3cm}
    \subfloat[Dataset B: Endoscopic images (ex-vivo)]{
    \label{fig:datasetb}\includegraphics[width=0.33\linewidth]{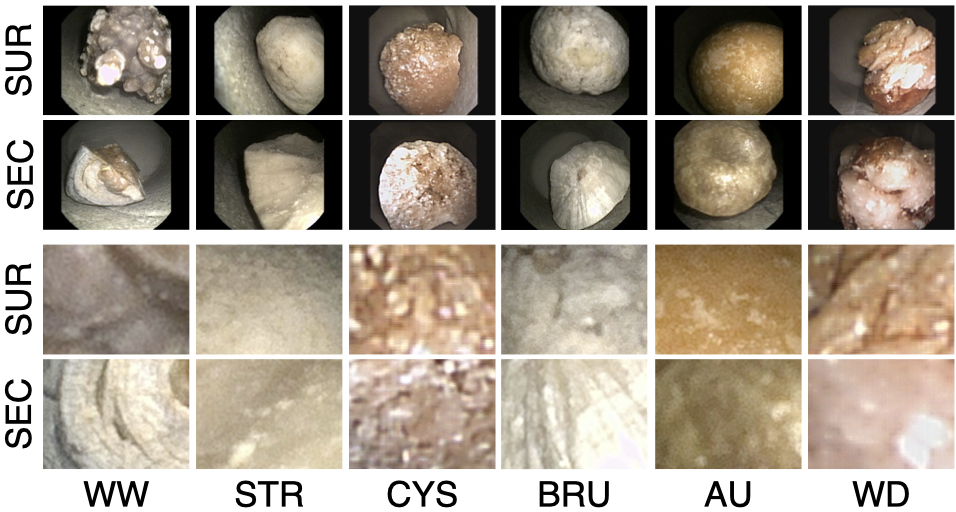}}
    \vspace{-0.1cm}
    \caption{Examples of ex-vivo kidney stone images acquired with (a) a CCD camera \cite{corrales2021classification} and  (b) an endoscope \cite{el2022evaluation}. SEC and SUR stand for section and surface views, respectively. The class types (WW, STR, CYS, etc.) are defined in Table \ref{tab:dataset}.}
    \label{fig:dataset}
    \end{figure*}

Transfer Learning (TL) is used when features learned from a given domain (or class of images) can bring appropriate knowledge to another domain for which the available image set is too small to train a large model from scratch \cite{raghu2019transfusion, wen2020effective}. 
In the context of ureteroscopy, a large dataset of in-vivo images is currently not available and collecting such a large database of endoscopic images during ureteroscopies is a long term work. However, in the context of this work, images of ex-vivo kidney stone fragments (acquired with standard CCD cameras) are available. Due to their similarity in color, texture and morphological features, TL can be used to distill knowledge from CCD-camera images
%the ex-vivo dataset (acquired with classical CCD cameras) 
into the final classifier of the images acquired with endoscopes.

Based on this idea, a two-step TL model to classify six types of kidney stones is proposed. The model uses a homogeneous, as well as a heterogeneous TL phase on a ResNet50 architecture pre-trained with the ImageNet dataset. To validate our proposal, the approach transfers knowledge learned on a small set of images acquired with classical CCD cameras to a final model that classifies ex-vivo endoscopic images. 
%The results obtained in this work improve those reported in the state of the art (up to 10\% in surface and section views measured with accuracy) for six classes of kidney stones in endoscopic images.

The rest of this paper is organized as follows. Section II provides an overview of the ex-vivo datasets, namely the CCD (digital) camera image and endoscopic image sets. Section II also presents the two-step TL approach. Section III compares the performances of the two-step TL approach with those of the methods of the literature. Finally, Section IV concludes this contribution and proposes perspectives.

\section{Materials and Methods} 
\subsection{Datasets} % Francisco
\label{datasets}
\vspace{-0.1cm}

Two kidney stone datasets were used in our experiments (see Table \ref{tab:dataset}): images were acquired with standard CCD cameras, and endoscopic images were captured with an ureteroscope. The dataset's main characteristics are described below.

% Michel Daudon:
\textbf{Dataset A, \cite{corrales2021classification}}.  This ex-vivo dataset of 366 CCD camera images (see the two 
upper lines in  Fig. \ref{fig:dataseta}) is split in 209 surface and 157 section images, and  contains six different stone types sorted by sub-types denoted by WW (Whewellite, sub-type Ia), CAR (Carbapatite, IVa), CAR2 (Carbapatite, IVa2), STR (Struvite, IVc), BRU (Brushite, IVd) and CYS (Cystine, Va). 
The fragment images were acquired with a digital camera under controlled lighting conditions and  with a uniform background. 

% Jonathan El-Beze 
\textbf{Dataset B, \cite{el2022evaluation}}. The endoscopic dataset  consists of 409 images (see the two upper lines in Fig. \ref{fig:datasetb}).
This dataset includes 246 surface and section 163 images. 
Dataset B involves the same classes as dataset A, except that the carbatite stones (sub-types IVa1 and IVa2) are replaced by the weddelite (sub-type IIa) and uric acid (IIIa) classes. 
The images of dataset B were captured with an endoscope by placing kidney stone fragments in an environment simulating in a realistic way the shape and color or ureters (for more details see \cite{el2022evaluation}). These images are visually close to in-vivo images since the fragments were acquired with an ureteroscope and by simulating a quite realistic the clinical in-vivo scenes.

\begin{table}[t!]
\centering
\caption{Description of the two ex-vivo datasets.}
\vspace{-0.1cm}
\label{tab:dataset}
%\scalebox{0.92}{%
\begin{tabular}{@{}cccccc@{}} 
\multicolumn{6}{c}{\textbf{Dataset A (M. Corrales et al. \cite{corrales2021classification})}} \\ \midrule  \vspace{-0.05cm}
Subtype & Main component & Key & Surface & Section & Mixed \\ \midrule  \vspace{-0.05cm}
Ia & Whewellite & WW & 50 & 74 & 124 \\  \vspace{-0.05cm}
IVa1 & Carbapatite & CAR & 18 & 18 & 36 \\  \vspace{-0.05cm}
IVa2 & Carbapatite & CAR2 & 36 & 18 & 54 \\  \vspace{-0.05cm}
IVc & Struvite & STR & 25 & 19 & 44 \\  \vspace{-0.05cm}
IVd & Brushite & BRU & 43 & 17 & 60 \\  \vspace{-0.05cm}
Va & Cystine & CYS & 37 & 11 & 48 \\ \cmidrule(l){3-6}  \vspace{-0.05cm}
 &  & TOTAL & 209 & 157 & 366 \\ \bottomrule   \vspace{-0.05cm}
 &  &  &  &  &  \\  
\multicolumn{6}{c}{\textbf{Dataset B (J. El-Beze et al. \cite{el2022evaluation})}} \\ \midrule  \vspace{-0.05cm}
Subtype & Main component & Key & Surface & Section & Mixed \\ \midrule  \vspace{-0.05cm}
Ia & Whewellite & WW & 62 & 25 & 87 \\  \vspace{-0.05cm}
IIa & Weddellite & WD & 13 & 12 & 25 \\  \vspace{-0.05cm}
IIIa & Acide Urique & AU & 58 & 50 & 108 \\  \vspace{-0.05cm}
IVc & Struvite & STR & 43 & 24 & 67 \\  \vspace{-0.05cm}
IVd & Brushite & BRU & 23 & 4 & 27 \\  \vspace{-0.05cm}
Va & Cystine & CYS & 47 & 48 & 95 \\ \cmidrule(l){3-6}   \vspace{-0.05cm}
 &  & TOTAL & 246 & 163 & 409 \\ \bottomrule  \vspace{-0.05cm}
\end{tabular}
%}
\end{table}

%\vspace{-0.5cm}

Due to the small size of the two datasets, and due to their class imbalance, patches were extracted from the images to increase and balance the number of training and testing samples. The two last lines of Figs. \ref{fig:dataseta} and \ref{fig:datasetb} show such patches.
As demonstrated in previous works \cite{lopez2021assessing, martinez2020towards, ochoa2022vivo}, the use of square patches of appropriate size allows to capture sufficient color and texture information for the classification. In addition, the use of patches instead of full surface and section images allows for augmenting and balancing the datasets. According to \cite{ochoa2022vivo}, a patch size of $256\times256$ pixels was chosen for the A and B datasets. A patch overlap of at most 20 pixels is set to avoid redundant information inside the image of a same kidney stone fragment. 
A total of 12,000 patches were extracted, both for dataset A and dataset B. % (1,000 patches per kidney stone type and view).  

The patches of each dataset were ``whitened" using the mean $m_i$  and standard deviation $\sigma_i$ of the color values $I_i^w$ for each RGB channel with  $I_i^w = (I_i – m_i)/\sigma_i$, with i = R, G, B.  %The strategy for partitioning the training and testing set was $80\%$ and $20\%$, respectively. 
To avoid data leakage in the datasets, a random, non-repeating dataset partitioning strategy was used (in contrast to previous works that used repeated images in both sets). $80\%$ of the data was used for training and $20\%$ for testing.

\subsection{Two-step transfer learning}
\label{transfer}
%\vspace{-0.1cm}

\begin{figure}[t]
\centering
\includegraphics[width=0.975 \linewidth]{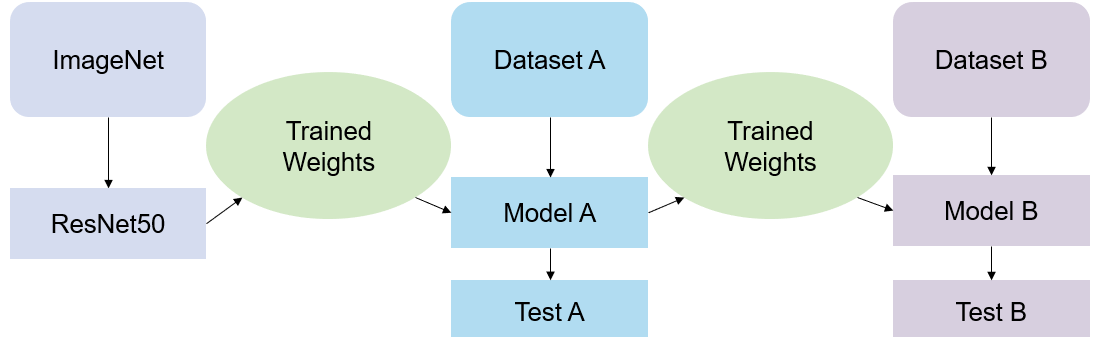}
\vspace{-0.1cm}
\caption{Two-step TL workflow. Model A was first initialized with the weights of a ResNet50 network pre-trained with ImageNet, and then fine-tuned with Dataset A. Next, Model B starts with the weights learned from Model A and is finally fine-tuned with Dataset B.}
\label{fig:method}
\end{figure}

%  ALREADY WRITEN  : please avoid repetitions
% Transfer Learning (TL) is a technique whose aim is to use the acquired knowledge of a model trained for a specific domain and apply that model to another domain, relying on the process of domain adaptation \cite{zhuang2020comprehensive}. 

Depending on the difference between two image domains, TL can be classified into homogeneous and heterogeneous transfer techniques. Homogeneous TL (HoTL) is applicable when specific datasets relating to a particular domain are available, even if the later is only ``similar'' to the images of the target domain (i.e., the dataset from which the knowledge is transferred does not exactly match the target dataset, but carries similar information). On the other hand, heterogeneous  TL (HeTL) is the case where the datasets of the source and target domains differ \cite{zhuang2020comprehensive, raghu2019transfusion}. Furthermore, when a reduced amount of training data is available, it is recommended to initialize the weights of architectures with pre-trained values rather than random values (i.e., TL from the scratch) \cite{raghu2019transfusion}. 
Thus, in the first step of the proposed two-step strategy, a large dataset (ImageNet) is used to transfer knowledge into a network (ResNet50) which is fine-tuned by the smaller kidney stone image set acquired under controlled acquisition conditions (dataset A). After this HeTL step, an HoTL is used, this second step exploiting dataset B including endoscopic images close to dataset A, but with more image quality variability as really encountered in ureteroscopy. 
%boost the capabilities of a given DL model for classifying kidney stones images from a endoscopic data distribution.

% Transfer 1
\textbf{First TL step: HeTL.}
%\subsection{First step of Transfer Learning}
Figure \ref{fig:method} shows the workflow of the two-step TL process. The first step is the heterogeneous phase in which the weights are initialized. To do so, a ResNet50 architecture was pre-trained with ImageNet  \cite{deng2009imagenet} and used to train a model able to classify the six types of kidney stones from dataset A (see Table \ref{tab:dataset}). 

In this  HeTL step, Gaussian blur and geometrical transformations are only applied to the training images with the aim of preparing the model for dataset B. A batch size of 24 was used along with a SGD optimizer with a learning rate of 0.001 and momentum of 0.9. Fully connected layers with 768,256,128 and 6 neurons were added with batch-normalization, ReLU activation function, and a dropout of 0.5.

% Transfer 2
%\subsection{Second step of Transfer Learning}
\textbf{Second TL-step: HoTL.}
The homogeneous learning occurs in the second step of the TL-process. It consists in transferring the knowledge (weights) of the trained model from the HeTL into dataset B to differentiate between the six types of kidney stones that are in this dataset (see Table \ref{tab:dataset}). 
% DO NOT UNDERSTAND THIS : Both steps were addressed to the fine-tuning approach, the architecture initialized with pre-trained weights and then those weights are re-trained on the new domain of the required task.?? 
%\textcolor{red}{
The initial weights of model B are those after the fine-tuning of model A with dataset A, model B being finally fine-tuned with dataset B.
%}. 
The purpose of this approach is to improve the generalization performance of model B and facilitate the extraction of robust features \cite{yosinski2014transferable}. Here, only geometric transformations were applied to the patches, since the image quality variability in dataset B is naturally high, while this variability is   limited  in dataset A. Moreover, 30 epochs were also executed with a SGD optimizer, but with a larger learning rate of 0.01 since it was expected that the model had less to learn. However, fully connected layers were not added since the idea was to use the trained model without further modifications to the architecture. 

\begin{table*}[]
\centering
\caption{Mean $\pm$ standard deviation determined for each metric quantifying the results for each patch type set  (fragment surface patches, section patches, and both patch types mixed) and for various TL-strategies  after 5 executions. Accuracy, Precision, Recall, and F1-Score were used to measure over six classes the performance of the models for each  TL-strategy.}
\label{tab:results1}
\vspace{-0.1cm}
\begin{tabular}{@{}cccccccl@{}}
%\toprule
Patch type & TL strategy & Accuracy & Precision & Recall & F1-Score & Dataset & Training details \\ \midrule \vspace{-0.025cm}
\multirow{5}{*}{Surface} & No TL & 0.582$\pm$0.033 & 0.588$\pm$0.028 & 0.582$\pm$0.033 & 0.579$\pm$0.028 & A & Baseline (no TL) trained on dataset A \\ \vspace{-0.025cm}
 & No TL & 0.702$\pm$0.012 & 0.718$\pm$0.010 & 0.702$\pm$0.012 & 0.701$\pm$0.008 & B & Baseline (no TL) trained on dataset B \\\vspace{-0.025cm}
 & HeTL only & 0.649$\pm$0.050 & 0.655$\pm$0.039 & 0.649$\pm$0.050 & 0.642$\pm$0.046 & A & Baseline + TL with ImageNet weights \\\vspace{-0.025cm}
& HeTL only & 0.820$\pm$0.033 & 0.833$\pm$0.029 & 0.820$\pm$0.033 & 0.818$\pm$0.032 & B & Baseline + TL with ImageNet weights \\\vspace{-0.025cm}
 & \textbf{HeTL$\,$+$\,$HoTL} & \textbf{0.832$\pm$0.012} & \textbf{0.845$\pm$0.012} & \textbf{0.832$\pm$0.012} & \textbf{0.829$\pm$0.012} & \textbf{B} & \textbf{Baseline$\,$+$\,$TL$\,$with ImageNet$\,$+TL$\,$dataset$\,$A } \\ \midrule
\multirow{5}{*}{Section} & No TL & 0.592$\pm$0.039 & 0.627$\pm$0.029 & 0.592$\pm$0.039 & 0.596$\pm$0.039 & A & Baseline (no TL) trained on dataset A \\ \vspace{-0.025cm}
 & No TL & 0.738$\pm$0.022 & 0.772$\pm$0.015 & 0.738$\pm$0.022 & 0.722$\pm$0.023 & B & Baseline (no TL) trained on dataset B \\ \vspace{-0.025cm}
 & HeTL only & 0.824$\pm$0.022 & 0.834$\pm$0.020 & 0.824$\pm$0.022 & 0.820$\pm$0.023 & A & Baseline + TL with ImageNet weights\\ \vspace{-0.025cm}
 & HeTL  only & 0.873$\pm$0.041 & 0.897$\pm$0.021 & 0.873$\pm$0.041 & 0.872$\pm$0.043 & B & Baseline + TL with ImageNet weights\\ \vspace{-0.025cm}
 & \textbf{HeTL$\,$+$\,$HoTL} & \textbf{0.904$\pm$0.048} & \textbf{0.915$\pm$0.037} & \textbf{0.904$\pm$0.048} & \textbf{0.903$\pm$0.050} & \textbf{B} & \textbf{Baseline$\,$+$\,$TL$\,$with$\,$ ImageNet$\,$+TL$\,$dataset$\,$A} \\ \midrule
\multirow{5}{*}{Mixed} & No TL & 0.594$\pm$0.021 & 0.610$\pm$0.023 & 0.594$\pm$0.021 & 0.596$\pm$0.020 & A & Baseline (no TL) trained on dataset A \\ \vspace{-0.025cm}
 & No TL & 0.760$\pm$0.024 & 0.773$\pm$0.029 & 0.760$\pm$0.024 & 0.752$\pm$0.024 & B & Baseline (no TL) trained on dataset B \\ \vspace{-0.025cm}
 & HeTL only & 0.800$\pm$0.013 & 0.809$\pm$0.013 & 0.800$\pm$0.013 & 0.797$\pm$0.013 & A & Baseline + TL with ImageNet weights\\ \vspace{-0.025cm}
 % 0.837±0.032	0.848±0.030	0.837±0.032	0.834±0.035
 & HeTL only & 0.837$\pm$0.032 & 0.848$\pm$0.030 & 0.837$\pm$0.032 &  0.834$\pm$0.035& B & Baseline + TL with ImageNet weights\\ \vspace{-0.025cm}
 & \textbf{HeTL$\,$+$\,$HoTL} & \textbf{0.856$\pm$0.001} & \textbf{0.868$\pm$0.002} & \textbf{0.856$\pm$0.001} & \textbf{0.854$\pm$0.001} & \textbf{B} & \textbf{Baseline$\,$+$\,$TL$\,$with$\,$ImageNet$\,$+TL $\,$dataset$\,$A} \\ \midrule \vspace{-0.025cm}
%\multicolumn{8}{l}{--- \textit{BASELINE: Model trained from scratch}}\\
%\multicolumn{8}{l}{* \textit{FINE-TUNING: Model pre-trained with ImageNet weights} }
\end{tabular}
\end{table*}

%\vspace{-0.2cm}

\begin{figure*}[]
\centering
\includegraphics[width=0.9 \linewidth]{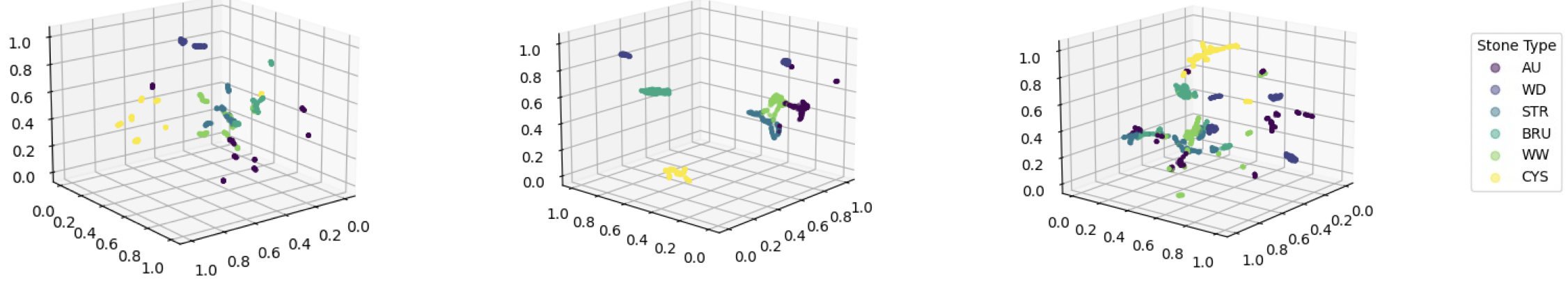}
\vspace{-0.2cm}
\caption{UMAP-ICN dimensionality feature reduction \cite{mendez2021finding}. From left to right: surface,  section, and mixed patch sets. The visualizations were generated in the second step of the  ``HeTL+HoTL'' strategy (see Fig. \ref{fig:method}).}
\label{fig:umap}
\end{figure*}

\section{Results and discussion} 
\vspace{-0.1cm}

Three different experiments were carried out to assess the performance of the two-step TL approach presented in Section \ref{transfer} using patch data described in Section \ref{datasets}. 
The ability of the two-step TL approach (see Fig. \ref{fig:method}) to predict kidney stone types on endoscopic and digital camera images was evaluated with surface (first experiment) and section patches (second experiment), these patch types being separately used. In the third experiment, based on a ``mixed'' dataset, the performance of the two-step TL approach was evaluated by simultaneously using surface and section patches. In previous works based on DL  \cite{lopez2021assessing, martinez2020towards, ochoa2022vivo} it has been reported that the combination of kidney stone patch types improves the classification process over models trained with only one patch type. Furthermore, mixing patch types closely simulates the way experts perform MCA and ESR \cite{corrales2021classification, daudon2016comprehensive}. 
%\textcolor{blue}{
The results of our experiments are summarized in Table \ref{tab:results1} and discussed below.%}
%By combining the surface and section views, the total number of patches used increases to 24,000, maintaining the ratio of 80\% and 20\%, for training and testing purposes respectively.
%In Table \ref{tab:results1} the well-known metrics (Accuracy, Precision, Recall, and F1-Score) are determined to measure the performance of the model for each view and step.

\subsection{Two step TL-approach results}
\textbf{Experiment 1.}
% Surface
From the surface patch results, it can be seen that the weights heterogeneously transferred 
%(different distributions)
from ResNet50 
%(pre-trained with ImageNet) 
to model A 
%(trained on microscopic images) 
(the ``HeTL only'' strategy applied on dataset A, see Table \ref{tab:results1}) led to an accuracy of 0.649$\pm$0.050) is useful to avoid training the model from scratch (accuracy of 0.582$\pm$0.033 for the ``No TL'' strategy applied on dataset A). 
Although the performance of ``HeTL only'' remains low, 
it is also noticeable that the two-step TL strategy  (``HeTL+HoTL'') of Fig. \ref{fig:method} 
improves significantly the identification  performance, obtaining an overall accuracy over 6 classes of 0.832$\pm$0.012 (increase of 18\% compared the ``HeTL only'' strategy with dataset A). The accuracy increase from ``HeTL only''  to the  ``HeTL+HoTL'' strategy is due to the similarity of the colors and textures of kidney stone fragments in databases A and B.

\textbf{Experiment 2.}
% Section
For section patches, the results of ``HeTL only'' applied on dataset A are promising (accuracy of  0.824$\pm$0.022). This high performance was reached due to the rich textural information in section patches which is not present in surface images. 
%. 
An accuracy of 0.904$\pm$0.048 was obtained for ``HeTL+HoTL'' applied on dataset  B.
Although the 8\% increase from  ``HeTL only'' to the ``HeTL+HoTL'' strategy was smaller for section data as for surface patches, this accuracy was the highest one in all three experiments.

% Mixed
\textbf{Experiment 3.} The ``HeTL'' strategy applied on mixed patches of dataset A, led to an accuracy of 0.8$\pm$0.013. This performance in increased by 5\% by the ``HeTL+HoTL'' strategy which has an accuracy of 0.856$\pm$0.001. In previous contributions it has  been reported that the simulatneous use of surface and section led to the highest performance.  This observation is not confirmed here, since the best results were obtained for section patches in experiment 2. 

%In general, the results obtained for the surface, section and mixed patch sets improve the results obtained from scratch, in the same way as using weights trained with ImageNet. 
Table \ref{tab:results1} shows that, in comparison to a learning from scratch,  all  TL-strategies improve the values of all four performance criteria, whatever the dataset.  
The UMAP-ICN visualisation \cite{mendez2021finding} of Fig. \ref{fig:umap} represents the features extracted in the last step of the ``HeTL+HoTL'' strategy. It is visible that for all patch types,  the inter-class distance is high and the intra-class distance is weak.

%\vspace{-0.1cm}

\subsection{Comparison with the state-of-the-art}
% SOTA comparative
Table \ref{tab:comparison} details the performance of reference DL-based methods used to identify the type of kidney stones using endoscopic image patches. These methods were \cite{estrade2022towards, black2020deep, martinez2020towards} were all implemented and evaluated on dataset B (endoscopic dataset), and compared to the TL-approach described in this contribution.
% SOTA ???? 
The results demonstrate that the two-step TL model outperforms the solutions described in \cite{estrade2022towards, black2020deep, martinez2020towards} in terms of accuracy. Thus, DL-strategies involving a pre-training with a general database, followed by a first tuning  with a specific database, and ending with a final tuning with the target database can effectively lead to an improved performance on different data distributions without the need of a large amount of data.

\begin{table}[t!]
\centering
\caption{Comparison of the performance of various kidney stone identification methods. The value of the accuracy over all classes was determined with dataset B for all methods.}
\vspace{-0.1cm}
%For more details of the dataset used see Sec. \ref{datasets}.}
\label{tab:comparison}
\begin{tabular}{@{}cccc@{}}
\toprule
Method & Surface & Section & Mixed \\ \midrule  \vspace{-0.025cm}
Martinez, et al. \cite{martinez2020towards} & 0.562$\pm$0.233 & 0.466$\pm$0.128 & 0.527$\pm$0.189 \\ \vspace{-0.025cm}
Black, et al. \cite{black2020deep} & 0.735$\pm$0.190 & 0.762$\pm$0.185 & 0.801$\pm$0.138 \\ \vspace{-0.025cm}
Estrade, et al. \cite{estrade2022towards}  & 0.737$\pm$0.179 & 0.788$\pm$0.106 & 0.701$\pm$0.223 \\ \vspace{-0.025cm}
\textbf{This contribution} & \textbf{0.832$\pm$0.012} & \textbf{0.904$\pm$0.048} & \textbf{0.856$\pm$0.001} \\ \bottomrule \vspace{-0.025cm}
\end{tabular}
\end{table}

\vspace{-0.3cm}

\section{Conclusion and perspective}
\vspace{-0.11cm}

It was demonstrated that it is possible to classify six different types of kidney stones using a small datasets of endoscopic images, the strategy being first to pre-learn the model with images acquired under controlled acquisition conditions (CCD camera) and then to exploit a fine tuning of the model using images captured in conditions simulating in a realistic way an ureteroscopy. 
This study confirms that it is easier for a neural network to adjust the weights learned on similar distributions and adapt them to a multiple class task. 
It is desirable that models of this type should be adapted to identify kidney stones using the complete endoscopic images instead of patches. We believe that the proposed approach will facilitate training on whole-image models when the datasets are reduced in the number of images.

\section*{Acknowledgments}
The authors wish to thank the AI Hub and the CIIOT at Tecnologico de Monterrey for their support for carrying the experiments reported in this paper in their NVIDIA's DGX computer.

We also wish thank the Verano de la Investigación Cientifica (VICI) Delfin program for assisting Juan Pablo Betancur-Rengifo with a mobility grant, and CONACYT for the doctoral scholarship for Francisco Lopez-Tiro at Tecnologico de Monterrey, and Ivan Reyes-Amezcua at CINVESTAV.

\section*{Compliance with ethical approval}
The images were captured in medical procedures following the ethical principles outlined in the Helsinki Declaration of 1975, as revised in 2000, with the consent of the patients.

\end{document}